\documentclass[preprint,authoryear,11pt]{elsarticle}

\usepackage[titletoc,title]{appendix}
\usepackage{multirow}
\usepackage{graphics}
\usepackage{booktabs}
\usepackage{array}
\usepackage[labelfont=bf]{caption}
\usepackage{amsmath}

\usepackage[outdir=./]{epstopdf}
\usepackage{array}
\usepackage{commath}
\usepackage[margin=0.9in]{geometry}

\bibpunct{[}{]}{;}{a}{,}{,}

\makeatletter
\def\ps@pprintTitle{%
  \let\@oddhead\@empty
  \let\@evenhead\@empty
  \def\@oddfoot{\reset@font\hfil\thepage\hfil}
  \let\@evenfoot\@oddfoot
}
\makeatother

\begin{document}

\begin{frontmatter}

\title{Autonomous CRM Control via CLV Approximation with\\ Deep Reinforcement Learning in\\ Discrete and Continuous Action Space}
\author{Yegor Tkachenko}
\address{Stanford University\\ \smallskip April 8, 2015}
\ead{yegor@stanford.edu}

\begin{abstract}
The paper outlines a framework for autonomous control of a CRM (customer relationship management) system. First, it explores how a modified version of the widely accepted Recency-Frequency-Monetary Value system of metrics can be used to define the state space of clients or donors. Second, it describes a procedure to determine the optimal direct marketing action in discrete and continuous action space for the given individual, based on his position in the state space. The procedure involves the use of model-free Q-learning to train a deep neural network that relates a client's position in the state space to rewards associated with possible marketing actions. The estimated value function over the client state space can be interpreted as customer lifetime value, and thus allows for a quick plug-in estimation of CLV for a given client. Experimental results are presented, based on KDD Cup 1998 mailing dataset of donation solicitations.
\end{abstract}


\end{frontmatter}

\section{Introduction}

Optimization of the direct marketing strategy with respect to each considered target individual is a pervasive challenge in marketing. Even with growing amounts of data, the models that would allow marketers to reliably determine the best direct marketing actions are rare, and direct marketing decisions are frequently based on guesswork and intuition rather than data. 

Fortunately, due to recent advances in the methodology of reinforcement learning and autonomous planning as well as due to our improving understanding of what metrics are key in predicting the client's response to direct marketing, systems for automated direct marketing planning and control are slowly becoming a reality.

In direct marketing, a company typically would like to know how it needs to engage with each client from its database to maximize the transaction rate during each subsequent period. The company would usually be able to engage with clients in several distinct ways (action space). The company would also possess some information indicative of what action would maximize the long-term gain from the given client.

The problem as presented yields itself well to Markov Decision Process formulation, where metrics the company possesses characterize the state of the client, and the probability and amount of transaction vary based on the client's state and the company's action. The objective is to determine which action, given the client's state, would maximize the cumulative (discounted) gain from the client. As an example of Markovian approach to marketing, see \citep{montoya2010dynamic}.

Unfortunately, coming up with a succinct and simultaneously rich characterization of the client is challenging. When such characterization is obtained, the state space often turns out to be extremely large, rendering model-based algortihms for solving MDPs unusable.

The rest of the paper is structured as follows. 

First, we explore how the state space of a customer can be summarized within a modified version of Recency-Frequency-Monetary Value framework.
Second, we discuss how model-free Q-learning can be used to train a deep neural network (also known as deep Q-network or DQN) based on historical information about the company's actions, the client's state transitions and response, and how the model's output can be interpreted as customer lifetime value of the given customer.
Third, we introduce a way to incorporate continuous action space into the deep Q-learning framework, allowing a deep neural net to learn value function over both discrete and continuous action spaces (mixed action space).
Finally, we train the model on mailing data from 1998 KDD Cup competition, showing that the proposed approach can lead to a significant boost in clients' response rates and donation amounts, and discuss how the methodology can be translated into a fully autonomous CRM system.

\section{RFM-I Customer Value Framework}

In marketing literature it is typical to summarize the state of a client using RFM framework - recency, frequency and monetary value. Recency is the number of periods since the client last transacted (note that \textit{recency} - the conventional name for this metric - is misleading - \textit{greater} recency implies \textit{smaller} value of the metric). Frequency is the number of transactions associated with the client during the observation window. Monetary value is the mean monetary value of the client's transactions.

As argued in \citep{cba,clv}, the RFM framework, while based on easily accessible data, succintly summarizes a wealth of information about the client. For example, if the client has been transacting very frequently, but recently has remained inactive for a long period of time, we would quickly begin to suspect that something happened and would think that the likelihood of the client's continuing to transact further is low. At the same time, we would not be concerned about similar period of silence from an individual who does not transact frequently. 

We choose to complement RFM framework with two additional metrics - frequency and recency of past marketing interactions (I), which would allow us to keep in check the potential deterring effect of marketing engagements that are too frequent or too concentrated in time.

Together, RFM-I metrics can serve as rich descriptors of the customer's state space based on data that is easily accessible in direct marketing settings. However, if more information on clients is available, one should feel free to incorporate it in the modeling process described below, which could accomodate thousands of variables. We now proceed to discuss how reinforcement learning can be used to build a connection between RFM-I (or potentially other metrics) and the expected reward (i.e. gain from the client).

\section{Deep Q-Learning}

The starting point of our analysis is the idea that the process of company's interaction with its customer base can be represented as an MDP. At each time-step the company selects an action $a_t$ from the set of available actions, $A = \{1, . . . , K \}$ (e.g. mailing types).
The company observes the RFM-I state of the client. Once the action has been selected, the company observes the reward (e.g. whether the client responds to marketing action), as well as the client's new RMF-I state.

Company's goal is to select such action that would maximize future rewards for each client. One approach to select such actions would be to use traditional model-based policy-search methods. However, these require explicit calculation of state transition probabilities and are of little use, as RFM-I state space can be very large.

An alternative is to use a model-free reinforcement learning method, e.g. Q-learning, which can learn from observed samples directly, without a need to construct the model of the environment. This is our preferred approach.

Following \citep{dql,hlcdrl,MK}, we assume that future rewards are discounted by a factor of $\gamma$ per time-step. We will have to estimate the optimal action-value function $Q^*(s, a)$, which is defined as the maximum expected return achievable, after seeing the client's state $s$ and then taking some action $a$.

The optimal action-value function obeys the Bellman equation. Intuitively, if the optimal value of the state $s'$ at the next time-step was known for all possible actions, then the optimal strategy is to select the action $a'$ maximizing the expected value of $r+\gamma Q^*(s',a')$.

\begin{equation}
Q^*(s,a) = R(s,a) + \gamma\sum_{s'}T(s'|s,a)\max_{a'}Q^*(s',a')
\end{equation}

Reinforcement learning aims to estimate the value function by using Bellman equation as an iterative update of the following form:

\begin{equation}
Q_{i+1}(s,a) = Q_i(s,a) + \alpha(r + \gamma \max_{a'}Q_i(s',a') - Q_i(s,a)),
\end{equation}

where $Q_i$ will aproach $Q^*$ as $i$ goes to infinity ($\alpha$ is the learning rate).

While this algorithm would converge to the optimal action-value function, this approach is impractical because the action-value function is estimated separately for each state without generalization. As already mentioned, state space as described by RFM-I framework can be very large, and so this becomes a particular problem.

Instead, a function approximator can be used to estimate the action-value function, $Q(s, a; \theta) \approx Q^∗(s, a)$. The model we utilize is a deep neural net (where `deep' simply refers to multiple hidden layers of the net). It is trained with Q-learning, with mini-batch gradient descent to update the weights, by minimizing a sequence of loss functions $L_i(\theta_i)$ that changes at each iteration $i$:

\begin{equation}
L_i(\theta_i) = E_{s,a}[(y_i-Q(s, a; \theta_i))^2],
\end{equation}

where 
\begin{equation}
y_i = r + \gamma \max_{a'}Q(s',a';\theta_{i-1}).
\end{equation}

Once the network has been trained, we can estimate $Q^*(s, a)$ for a client in a given state and then extract the optimal action.

Reinforcement learning has been shown to be unstable or even to diverge when the Q-value function is approximated with non-linear functions (including neural networks) \citep{tsitsiklis1997analysis}.

This instability is rooted in the difference between assumptions made by supervised learning and reinforcement learning. Supervised learning assumes that data samples are independently drawn from a fixed underlying distribution, while reinforcement learning deals with sequences of correlated states sampled from a distribution that changes as the agent adopts new policies over time.

To mitigate such issues, deep Q-learning uses a mechanism called experience replay \citep{dql,hlcdrl}: the DQN is trained on episodes sampled uniformly from a replay memory, therefore smoothing the data distribution and removing correlations in the sequence of observations. 

We use historical records for learning in our experiments and are thus able to randomly sample previous transitions, smoothing the training distribution over many past behaviors. By doing this we effectively implement a version of experience replay, which allows us to mitigate the problems of correlated data.

Use of a separate network for generating the targets in the Q-learning update, coupled with only periodical updates of the action-value targets, is another trick that can be helpful in mitigating the issue, as most recently proposed in \citep{hlcdrl}. More precisely, every $x$ number of updates the network Q is cloned to obtain a target network Q*, which is then used for generating $\max_{a}Q^*(s',a')$ targets for the updates to Q.

\section{Deep Q-Network as Customer Lifetime Value Approximator}

If we consider Bellman equation (1) carefully, it is apparent that the estimated Q-value function on monetary reward gives us expected future discounted financial flow from the client in the given state for each action we could take at the current period, assuming we act optimally in all subsequent periods. This is precisely how we would define the expected customer lifetime value (CLV) from the current state of the client and on (sometimes also referred to as \textit{residual} CLV \citep{fader2012reconciling}), conditional on the actions taken with respect to the client.

Thus, when deciding which action to take within the deep Q-learning framework, we are in fact using CLV maximization as our guiding principle - which fully follows business rationale. This interpretation renders deep Q-learning an especially rich framework for business applications.

The proposed deep reinforcement learning approach compares favorably to other CLV modeling techniques in allowing us to incorporate marketing actions into the decision making process. Other CLV modeling techniques may offer greater interpretability due to a particular parametrization \citep{gupta2006modeling}, however, they often become mathematically unwieldy when state and action spaces grow. Moreover, other approaches may be at a disadvantage when value function is highly non-linear, whereas a multilayer neural network, by a universal approximation theorem \citep{hornik1991approximation}, can theoretically fit any function arbitrarily well, no matter how complex. The expressiveness and computational efficiency thus make deep Q-learning an attractive instrument for real-life autonomous direct marketing control.

\section{Continuous Action Space}

Typically, deep Q-network takes \abs{s} state variables as input and has \abs{a} output neurons - each calculating the value of the corresponding discrete action, given the input state \citep{dql,hlcdrl}. 
If the action space is continuous, two approaches are possible. First, one could discretize the continuous action space, increasing the number of output neurons. However, if the actions are characterized by multiple continuous variables, this could lead to either or both the loss of valuable information and an extremely large action space \citep{gasketto}.

Luckily, deep neural networks allow for another approach, where continuous variables characterizing the actions can enter as input of the neural net together with state space variables. This way the neural net approximates the value of each discrete action given (a) its continuous characteristics and (b) the state space.

One challenge with this approach is the need to calculate $\max_{a'}Q_i(s',a')$ over action space in the iterative Bellman updates or $\arg\max_{a'}Q_i(s',a')$ when extracting optimal policy. When action space is discrete, we simply choose the action that corresponds to the output neuron with the greatest value. In the case of continuous actions we need to first find the maximum value of each discrete action (output neuron) over the continuous action space, while holding the state space fixed. Then, as in the discrete case, we choose the action corresponding to the maximum value neuron.

While optimizing over the continuous action space may sound to be a cumbersome task, chain rule used in backpropagation algorithm allows us to easily calculate gradients of the Q-values on the continuous action space input variables. This enables us to use efficient gradient descent methods to estimate what continuous action variables maximize the expected value of the given discrete action.

This approach proves to be viable and constitutes a novel contribution of the paper.

\section{Experimental Results}

We use 1998 KDD Cup competition mailing data in the experiment \citep{kdd98}. It represents a record of donation-seeking mailings to the total of 95,412 clients during 23 distinct periods. The type of mailing changed from period to period. We observe 12 discrete possible actions (11 mailing types + inaction). Inaction corresponds to missing action indicator in the KDD Cup data set. The continuous aspect of the action is the number of the month when the mailing was sent (essentially a discrete value, but, as a toy example, we treat it as a continuous variable).

We extract from the available data the total of 22 * 95,412 state transitions. For each state transition we know RFM-I metrics (how recently the person donated last, how frequently he donated, his average donation amount, how recently and how many times we sent him the mail), discrete action taken by the company, a continuous aspect of the action and a reward observed (monetary value of the donation). A sample record looks as follows: $$[r_t, f_t, m_t, ir_t, if_t, a_t, acont_t, r_{t+1}, f_{t+1}, m_{t+1}, ir_{t+1}, if_{t+1}, reward]$$

We then use deep Q-learning to train a model as explained in the previous section, and extract the optimal policy for each of the starting states in the state transition chains we used in training the model.

We run comparison on policies derived from two different models: one trained on discrete actions only (ignoring the continuous action space during training), second trained on discrete \textit{and} continuous action space. When we extract the optimal policy in discrete \textit{and} continuous action space, we round the continuous action space variable(s) to enable matching with historical records for policy evaluation.

The network architecture is as follows: 2 hidden layers (40 and 15 neurons respectively with ReLU activation function of form $\max(0,x)$), 12 regression output neurons - one for each discrete action (including inaction - action 0), 5 or 6 variable inputs (5 RFM-I state variables + 1 continuous action variable), depending on whether we include the continuous action variable in the modeling process. For the discrete action model input space is limited to 5 state variables.

Each model is trained with mini-batch gradient descent (batches of 200 transitions each) using RMSProp algorithm \citep{RMSProp} for 100 epochs, where epoch marks a point when each sample in the training data set has been sampled once (training length of 100 epochs means that each observation contributed to gradient update of the neural network 100 times). We begin with 0.001 learning rate and use 0.99 decay rate (learning rate is multiplied by decay rate every epoch, allowing us to further fine-tune the learning process as we observe more samples). We use 0.9 discount rate. Additionally, as a way to facilitate convergence (discussed earlier), every $10,000$ iterations we clone the network Q to obtain a target network Q*, which we then use for generating $\max_{a}Q^*(s',a')$ components of updates to Q.

To evaluate the effectiveness of the resulting policy, we compare (a) the average response rate and donation amount in situations (groups of interactions between the company and its clients) when actions of the company were aligned with model's recommendations to (b) the average response rate and donation amount in those instances when company's actions deviated from the model's recommendations. 

We also compare the performance of the extracted policy to additional benchmarks: (a) reward from a random policy, (b) reward of always taking best-performing action - mailing type number 4 - which is also a policy selected by a simple regression model approximating donation amounts on RFM-I input state space, and (c) average reward achieved across the dataset.

During training, we select the best model by evaluating the resulting policy, as described above, on a hold-out validation set. The training set is 1.6 million randomly sampled transition tuples. The validation set is approximately 0.5 million transition tuples.

Note that while we use a validation set to pick the best model during training, the final evaluation is run on the full data set (to see how the policy is doing for the whole customer base). The obtained metrics follow almost exactly those calculated exclusively on the validation data set.

We find that the average response rate and monetary reward (donation amount) are significantly higher when the policy is adhered to - for both trained models. Table 1 presents the evaluation results.

However, the estimation process became much less stable with incorporation of continuous action space - possibly due to correlations between continuous action inputs and the network outputs - despite measures undertaken to facilitate convergence (e.g. experience replay and a lag in the model update, as described previously). In particular, the success of convergence would heavily depend on specific weight initializations of the network parameters. 

As a result, policy from the model trained on continuous and discrete action space achieved less of an improvement in the performance metrics, compared to policy from the simpler discrete-action-only model. 

Nevertheless, we show that training DQN directly on the continuous action space is a viable and effective approach that should be explored further. The issue with convergence indicates a need for further research on ways to ensure stability when training deep neural nets.

\begin{table*}
\huge
\caption{Response rates for groups of interactions between the company and the client, depending on whether the type of mailing sent to the client agreed with or contradicted the optimal policy, given the client's RFM-I state}
\renewcommand{\arraystretch}{2}
\centering
\resizebox{\columnwidth}{!}{%
\begin{tabular}{p{18cm}>{\centering\arraybackslash}p{12cm}>{\centering\arraybackslash}p{12cm}}
  \hline
&\multicolumn{1}{c}{Average response probability} & \multicolumn{1}{c}{Avergae donation amount (\$)} \\
\cmidrule(lr){1-3} 

\multicolumn{3}{l}{\textbf{Policy 2: DQN model trained on the discrete action space}}\\ \cmidrule(lr){1-3}
Marketing done as recommended by the model & 30.9\% & 4.9 \\
Marketing deviated from the model's recommendation & 11.8\% & 1.6 \\
   \hline
\multicolumn{3}{l}{\textbf{Policy 1: DQN model trained on the discrete and continuous action space}}\\ \cmidrule(lr){1-3}
Marketing done as recommended by the model & 30.5\% & 4.3 \\
Marketing deviated from the model's recommendation & 11.7\% & 1.6 \\
   \hline
\multicolumn{3}{l}{\textbf{Other benchmarks}}\\ \cmidrule(lr){1-3}
Random policy & 18.9\% & 2.6 \\
Only the top performing action (\# 4) & 24.4\% & 3.4 \\
Mean reward across the data set & 12.6\% & 1.8 \\
   \hline
   
\end{tabular}
}
\end{table*}

In presenting the results, we focus only on the better performing discrete action model. We summarize the derived policy in Table 2. Figure 1 shows the training progress.

\begin{table*}
\huge
\caption{Summary of recommended actions for the policy from the discrete action model}
\renewcommand{\arraystretch}{2}
\captionsetup{justification=centering}
\centering
\resizebox{\columnwidth}{!}{%
\begin{tabular}{cccccccc}
  \hline
\multicolumn{4}{c}{} & \multicolumn{4}{c}{Average by the client group} \\ \cline{5-8}
 Action Number & Mailing Description & Frequency & Proportion & Response Rate & Recency & Frequency & Monetary Value\\ \cline{1-8} 
\hline 

\textbf{2} & Simple mailings with labels & 27,946  & 1.3\%  & 12.8\% &  11.5	& 1.2	& 13.5 \\

\textbf{4} & Blank cards that fold into thirds with labels & 1,454,421  & 69.3\%  & 11.6\% &	4.4 &	2.1	& 12.4 \\

\textbf{7} &\textit{Thank you} printed with labels & 373,451 & 17.8\%	& 15.8\% &	2.3 &	0.6 &	2.4	 \\

\textbf{9} & Christmas cards with labels & 34,731  & 1.7\%  & 9.8\% & 9.7	& 1	& 9.4 \\

\textbf{10} & Mailings with labels and a notepad (a) & 36,379  & 1.7\%  & 3\% &  11.1  & 2.4	& 19 \\

\textbf{11} & Mailings with labels and a notepad (b) & 166,988  & 8.0\%  & 15.5\% & 2.2	& 1.6	& 15.3 \\

\hline
   
\multicolumn{8}{l}{*Actions recommended less than 1\% of the time have been omitted. See documents accompanying the data set for the full action list.} \\
\end{tabular}
}
\end{table*}

Finally, we explore how the output of the model can be examined visually. Figures 2-8 show the expected cumulative discounted reward associated with each action, given the particular dimension(s) of the client's state. Lines/surfaces in these plots correspond to actions. We would need a 5-dimensional plot to summarize the full decision surface learned by the neural network (5 state dimensions). This is challenging, and thus we resort to lower-dimensional plots. 

The plots are also curious in their own right in showing how the reward changes with RFM-I metrics. For example, it can be clearly seen that too many marketing interactions lead to a decrease in donations, and that long inactivity, as captured by recency metric, hints at decreased response likelihood for clients who transact frequently. The model was also able to detect increased reward associated with sending `thank you' note (action 7) when the client transacted very recently and decreased value of such action when the client has not transacted for some time. The nonlinear shape of the value function suggests that models simpler than DQN would have hard time fitting all the intricacies of CLV over the customer state space.

\section{Autonomy}

So far we have discussed the machinery necessary to estimate the long-term value associated with each action a company may take with respect to each client, based on the historical data that is available. Only a small step is needed to make the system autonomous - that is, capable of running on its own even in the cold start situations.

The idea is to enable the sytem to build its own database of observations for training through random exploration. In other words, as the system is launched and has no observation history, it takes an action at random with respect to each client with probability 1. Over time, as the model is trained on collected observations, the probability of taking a random action begins to decrease, whereas the probability of taking the action deemed optimal by the model goes up, as the model begins to exploit collected information. 
It is also important to keep random exploration on with some small probability even when a significant amount of history has been accumulated - to ensure the system can detect changes in effectiveness of actions.

Some additional modifications can make the system more effective. Below we provide a non-exhaustive list.

\begin{itemize}
  \item A custom policy could be defined for the system to pursue before enough observations have been accumulated.
  \item Some more sophisticated exploration strategies may be implemented - e.g. encouraging exploration of actions that have been relatively `underobserved'.
  \item Additional constraints may be imposed - e.g. not to take a particular action unless a condition has been met (not sending `thank you' note unless a customer has transacted recently) etc.
  \item Actions that are significantly underperforming may be dropped from the consideration pool.
\end{itemize}

As the system begins to accumulate observations, experience replay mechnanism kicks in. Observations are stored in replay memory and are then sampled randomly for mini-batch gradient descent updates. As noted earlier, this allows us to smooth the data distribution and mitigate the problems of correlated data. As the training proceeds, the quality of recomended policy improves.

It can thus be seen that training of the deep Q-network from scratch fits  well into the autonomous CRM control process.

Note that the described system is autonomous \textit{strictly given the specified action space} - i.e. a marketer needs to first determine the actions that can be taken by the system, which would typically involve the creative human design process (e.g. writing up the text of the mailed ad, designing the visuals etc.). While such pre-defined action space is a conventional prerequisite for many systems referred to as `autonomous' in the literature, devising a way for the system to come up with \textit{new} actions on its own is a very promising field of research and is a necessary step towards a system that is autonomous in the true sense of the word. This issue, however, is beyond the scope of this paper. 

\section{Conclusion}

We have shown that by combining RFM-I client space parametrization with deep Q-learning in planning the direct marketing actions, significant improvements in the client response rate and donation amounts can be achieved (over 50\% for the particular dataset we experimented with).

We also propose a novel method for incorporating continuous actions into the deep neural net training procedure. However, further advances in procedures for training deep neural nets are needed to ensure consistent convergence when training DQN directly on the continuous action space.

We note that the Q-value function approximated by the deep neural net on monetary reward data constitutes an estimate of action-dependent (residual) customer lifetime value (CLV), which is precisely the criterion we want to use in selecting the optimal marketing actions. CLV could also be a useful metric for other purposes, e.g. corporate valuation \citep{bauer2005customer}, segmentation etc. Thus, a trained deep neural net that works as a plug-in estimator of CLV for any client we may observe in the data can prove to be an extremely valuable tool in multiple business settings.

The results suggest that once possible interactions with clients have been defined and a record of such interactions has been accumulated, deep Q-learning on RFM-I metrics space constitutes an effective approach to direct marketing control. Moreover, as discussed, the system can be easily set up to build its own database of observations for training through random exploration, allowing it to work in cold start situations - i.e. without existing historical records, and thus rendering it autonomous.

It is also clear that there is space for more research in the field. For example, further improvements in techniques for training the deep neural nets are necessary to ensure stable convergence of the more complex DQN models. Another interesting challenge is the extension of the proposed approach to more complex domains - e.g. e-commerce with numerous product categories and a variety of actions that are specific to particular purchase histories of clients. Finally, devising algorithms that would allow the system to come up with new actions is an extremely promising field of research.

We conclude that, although in its infancy, deep reinforcement learning can serve as a foundational control methodology for autonomous CRM and other management systems. 

\begin{figure*}[p]
    \centering
    \includegraphics[width=9cm]{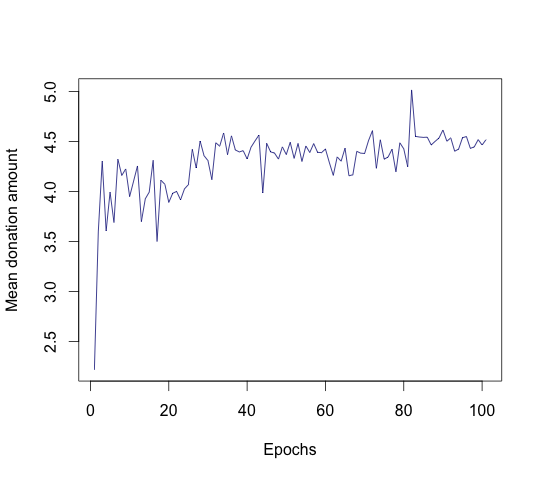}
    \caption{Training history for the discrete action model: average donation amount when marketing is aligned with the model's recommendations (values are calculated on the validation data set)}
    \includegraphics[width=9cm]{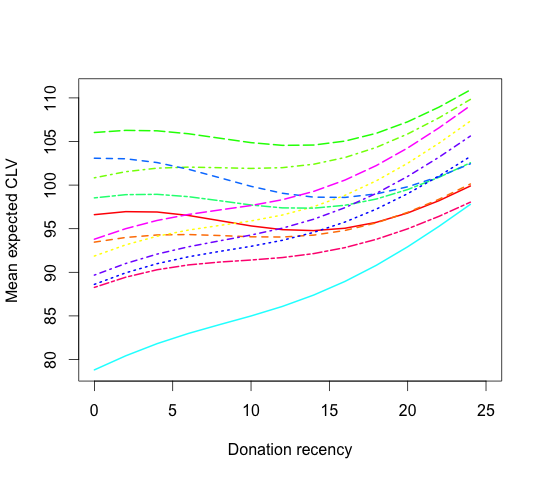}
    \includegraphics[trim = -5mm -35mm 0mm 0mm, clip, width=1.7cm]{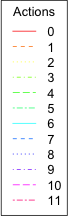}
    \caption{Expected cumulative discounted reward associated with each action, based on the recency of donations}
\end{figure*}

\begin{figure*}[p]
    \centering
    \includegraphics[width=9cm]{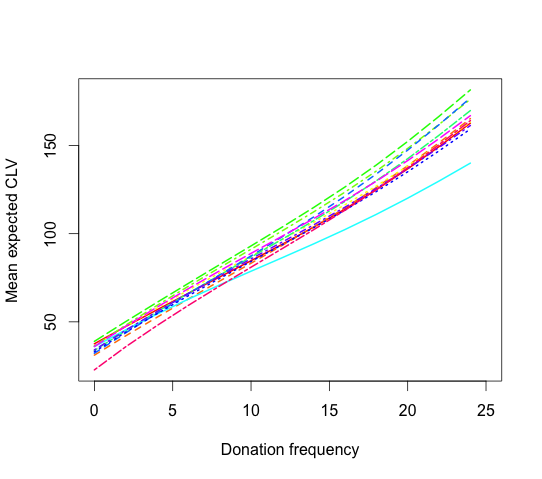}
    \includegraphics[trim = -5mm -35mm 0mm 0mm, clip, width=1.7cm]{legend.png}
    \caption{Expected cumulative discounted reward associated with each action, based on the frequency of donations}
    \includegraphics[width=9cm]{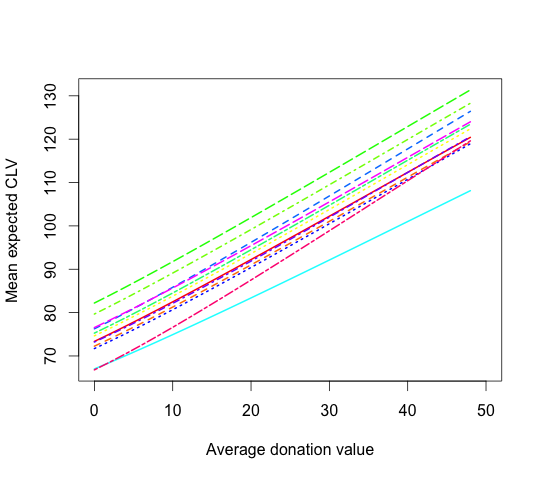}
    \includegraphics[trim = -5mm -35mm 0mm 0mm, clip, width=1.7cm]{legend.png}
    \caption{Expected cumulative discounted reward associated with each action, based on the client's average donation}
\end{figure*}

\begin{figure*}[p]
    \centering
    \includegraphics[width=9cm]{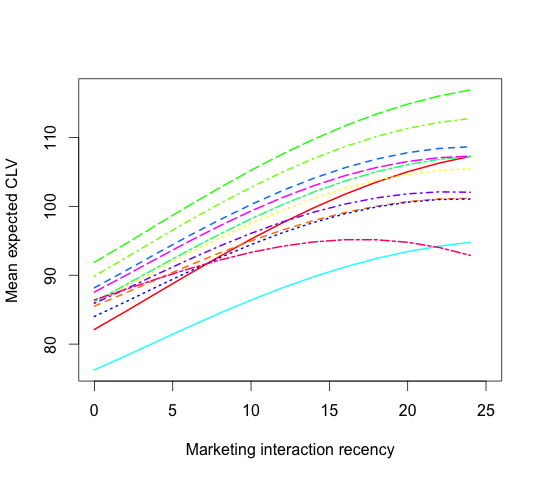}
    \includegraphics[trim = -5mm -35mm 0mm 0mm, clip, width=1.7cm]{legend.png}
    \caption{Expected cumulative discounted reward associated with each action, based on the recency of past marketing interactions}
    \includegraphics[width=9cm]{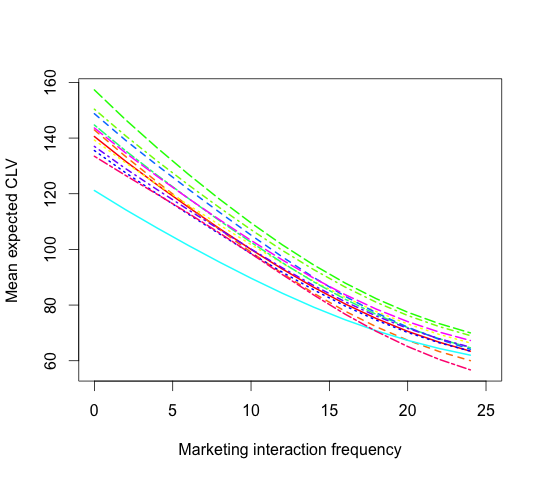}
    \includegraphics[trim = -5mm -35mm 0mm 0mm, clip, width=1.7cm]{legend.png}
    \caption{Expected cumulative discounted reward associated with each action, based on the number of past marketing interactions}
\end{figure*}

\begin{figure*}[p]
    \centering
    \includegraphics[width=9cm]{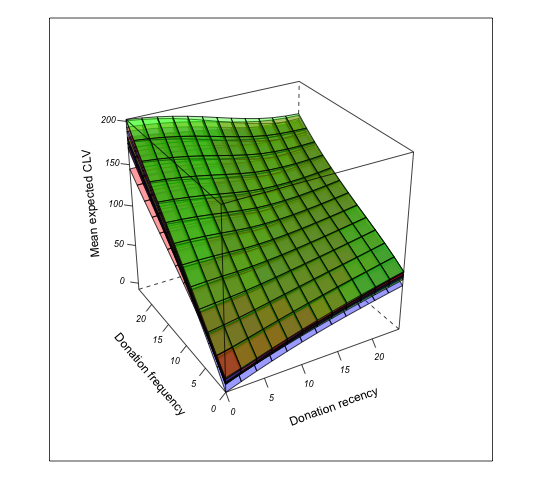}
    \caption{Expected cumulative discounted reward associated with each action, based on the client's position in Recency-Frequency space (each surface represents an action, dominant green layer corresponds to action 4)}
    \includegraphics[width=9cm]{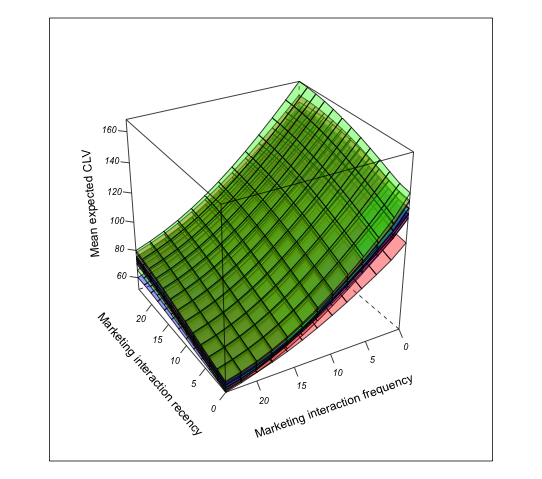}
    \caption{Expected cumulative discounted reward associated with each action, based on the client's position in Marketing Interaction Recency-Frequency space (each surface represents an action, dominant green layer corresponds to action 4)}
\end{figure*}

\newpage

\bibliographystyle{elsarticle-harv}

\end{document}